\pdfoutput=1

\documentclass[11pt]{article}

\usepackage[]{acl}
\usepackage{multirow}
\usepackage{graphicx}
\usepackage{booktabs}
\usepackage{tabularx}
\usepackage{CJKutf8}
\usepackage{amssymb}
\usepackage{amsmath}
\usepackage{tabu} 
\usepackage{hyperref}
\usepackage{breakurl}
\usepackage{times}
\usepackage{latexsym}
\usepackage{tikz}
\usepackage{pgfplots}
\usepackage[normalem]{ulem}
\usepackage{xcolor}
\definecolor{lightblue}{RGB}{79,173,234}
\usepackage{dashrule}
\usepackage{microtype}
\usepackage{tikz}
\usepackage{pgfplots}
\usepackage{CJKutf8}
\usepackage{tikz-dependency}
\usepackage{graphicx}
\usepackage{natbib}
\usepackage{tikz-qtree}
\usepgflibrary{arrows.meta}
\usepgfplotslibrary{fillbetween}
\usetikzlibrary{fit, calc, decorations.pathreplacing, positioning, shapes.geometric, backgrounds}
\usepackage{float}
\usepackage{caption}
\usepackage{subcaption}
\definecolor{brickred}{HTML}{b92622}
\definecolor{midnightblue}{HTML}{005c7f}
\definecolor{salmon}{HTML}{f1958d}
\definecolor{burntorange}{HTML}{f19249}
\definecolor{junglegreen}{HTML}{4dae9d}
\definecolor{forestgreen}{HTML}{499c5e}
\definecolor{pinegreen}{HTML}{3d8a75}
\definecolor{seagreen}{HTML}{6bc1a2}
\definecolor{limegreen}{HTML}{97c65a}

\usepackage{lipsum}

\newcommand\blfootnote[1]{%
  \begingroup
  \renewcommand\thefootnote{}\footnote{#1}%
  \addtocounter{footnote}{-1}%
  \endgroup
}

\usepackage[T1]{fontenc}

\usepackage[utf8]{inputenc}

\usepackage{microtype}
\usepackage{arydshln}

\title{CSynGEC: Incorporating Constituent-based Syntax for Grammatical Error Correction with a Tailored GEC-Oriented %
Parser
}
 
\author{Yue Zhang \and Zhenghua Li$^{*}$ \\
        Institute of Artificial Intelligence, School of Computer Science and Technology, \\
Soochow University, China;\\
\texttt{yzhang21@stu.suda.edu.cn}, \texttt{zhli13@suda.edu.cn}\\}

\begin{document}
\begin{CJK}{UTF8}{gkai}

\maketitle
\begin{abstract}
Recently, \citet{zhang2022syngec} propose a syntax-aware grammatical error correction (GEC) approach, named SynGEC, showing that incorporating tailored dependency-based syntax of the input sentence is quite beneficial to GEC. 
This work considers 
another mainstream syntax formalism, i.e.,  constituent-based syntax.
\textcolor{black}{
By drawing on the successful experience of SynGEC, 
we first propose an extended constituent-based syntax scheme to accommodate errors in ungrammatical sentences.
Then, we automatically obtain constituency trees of ungrammatical sentences to train a GEC-oriented constituency parser by using parallel GEC data as a pivot.
For syntax encoding, we employ the graph convolutional network (GCN).
}
Experimental results show that our method, named CSynGEC, yields substantial improvements over strong baselines. 
Moreover, we investigate the integration of constituent-based and dependency-based syntax for GEC in two ways: 
\textcolor{black}{
1) intra-model combination, which means using separate GCNs to encode both kinds of syntax for decoding in a single model; 2)inter-model combination, which means gathering and selecting edits predicted by different models to achieve final corrections. 
We find that the former method improves recall over using one standalone syntax formalism while the latter improves precision, and both lead to better F${_{0.5}}$ values. \blfootnote{$^*$ Corresponding author.}
}

\end{abstract}
\section{Introduction}

\textcolor{black}{
Grammatical error correction (GEC) aims to craft a grammatical and fluent version of an ungrammatical sentence with the intended meaning \citep{wang2021comprehensive}.
As an important yet challenging task in NLP, GEC lays foundations for many downstream applications like writing assistance and second language acquisition \cite{bryant2022survey}.
}

In recent progress, Transformer \cite{vaswani2017attention} emerges as the most powerful backbone for GEC \cite{rothe2021recipe,zhang-etal-2022-mucgec}. Despite its remarkable success, some recent works point out that Transformer can be further strengthened when integrated with syntactic information \cite{wan2021syntax,li2022syntax}.
\textcolor{black}{
However, there lies a challenge for syntax-enhanced GEC in that off-the-shelf parsers may not handle ungrammatical inputs well. To tackle this problem, \citet{zhang2022syngec} present the SynGEC approach. Specifically, they build a GEC-oriented dependency parser by extending the original syntax scheme and utilizing parallel GEC data to automatically generate trees of ungrammatical inputs for training. Such a tailored dependency parser helps SynGEC stand out from all syntax-enhanced GEC methods and achieve state-of-the-art (SOTA) results.
}

\begin{figure*}[th!]
  \begin{subfigure}[b]{0.2\textwidth}
    \centering
    \includegraphics[scale=0.15]{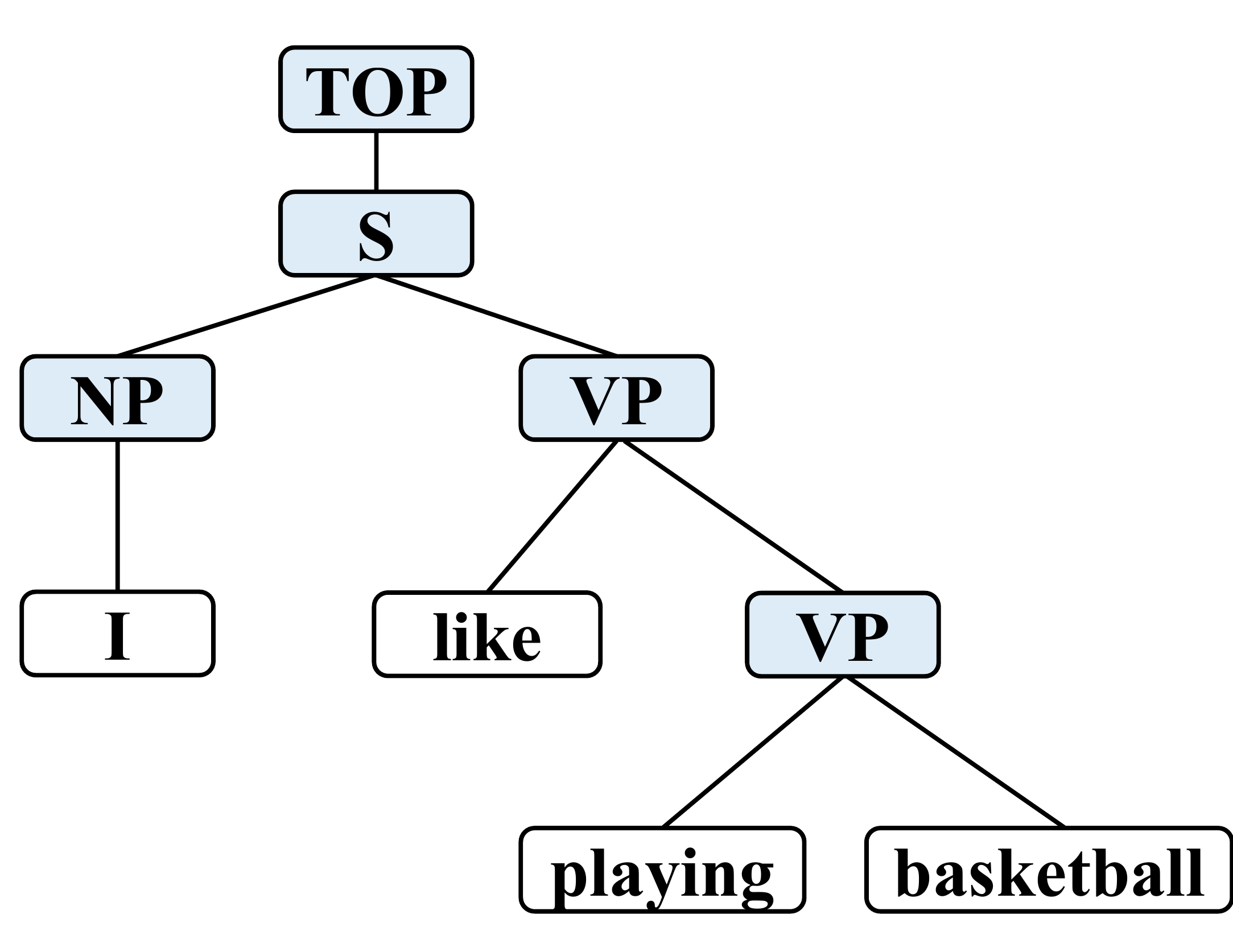}
    \caption{The correct sentence.}
    \label{fig:origin-sent}
  \end{subfigure}
  \hfill
   \begin{subfigure}[b]{0.2\textwidth}
    \centering
    \includegraphics[scale=0.15]{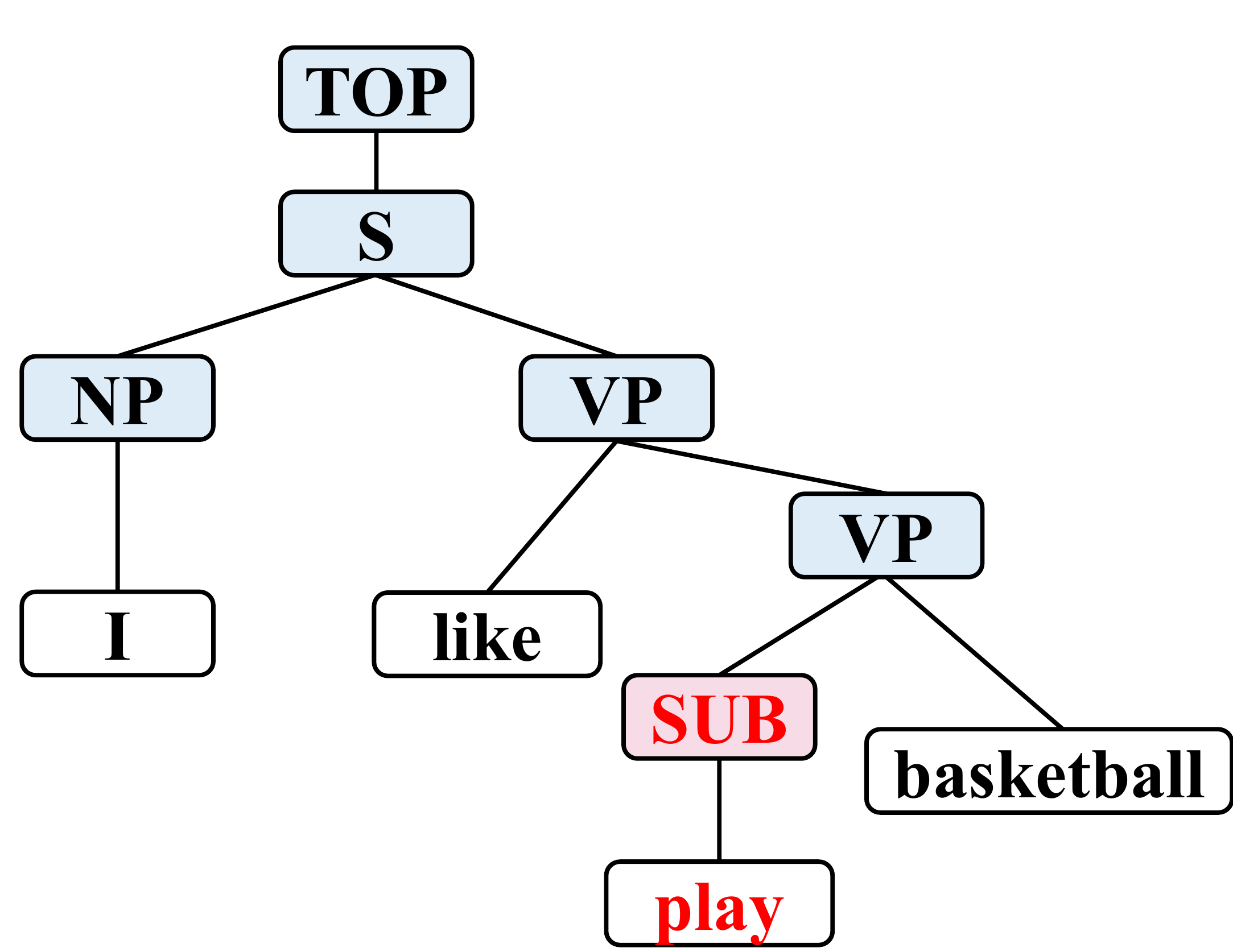}
    \caption{Substituted errors.}
    \label{fig:s-error}
  \end{subfigure}
  \hfill
  \hfill
   \begin{subfigure}[b]{0.2\textwidth}
    \centering
     \includegraphics[scale=0.15]{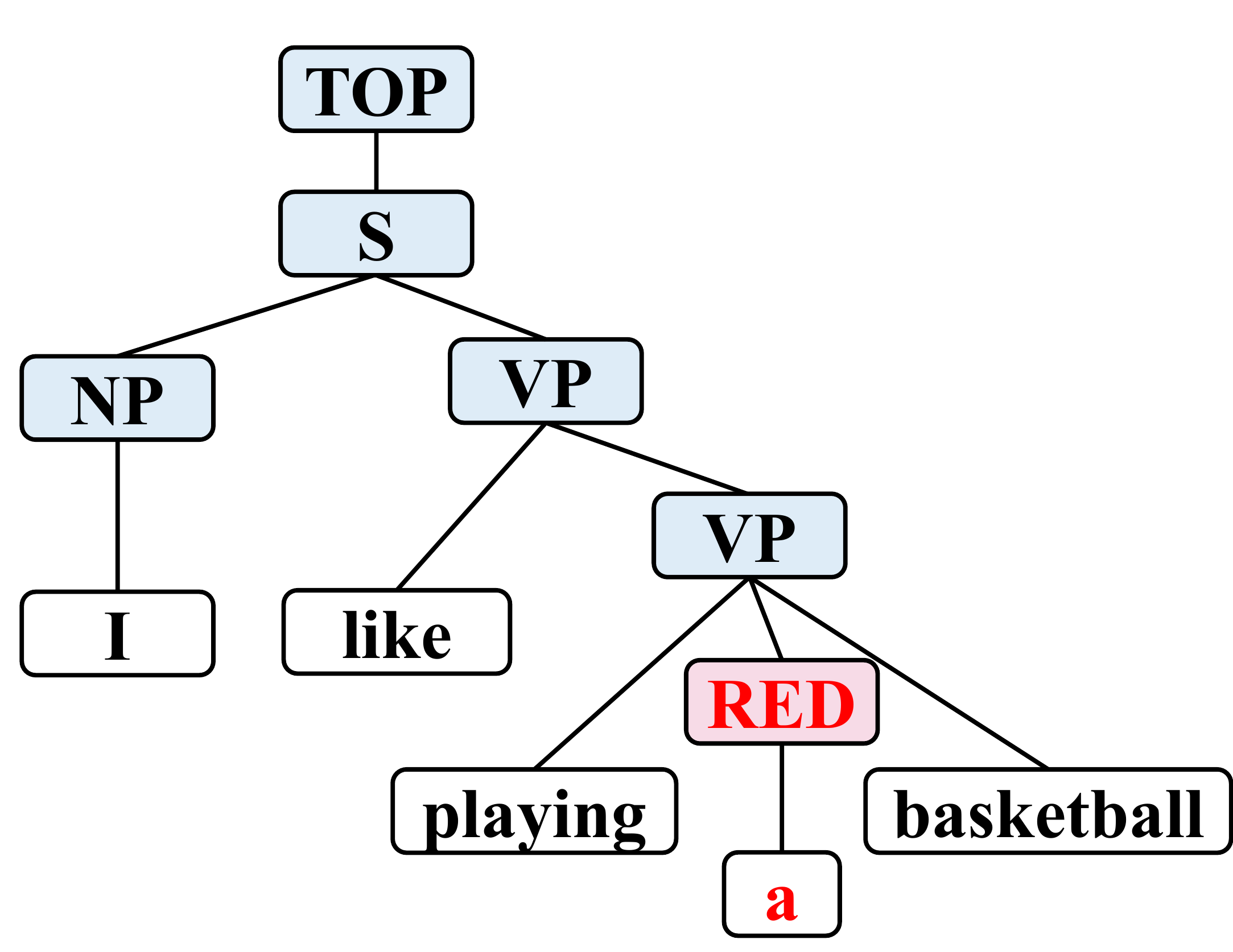}
    \caption{Redundant errors.}
    \label{fig:r-error}
  \end{subfigure}
  \hfill
  \hfill
   \begin{subfigure}[b]{0.2\textwidth}
    \centering
     \includegraphics[scale=0.15]{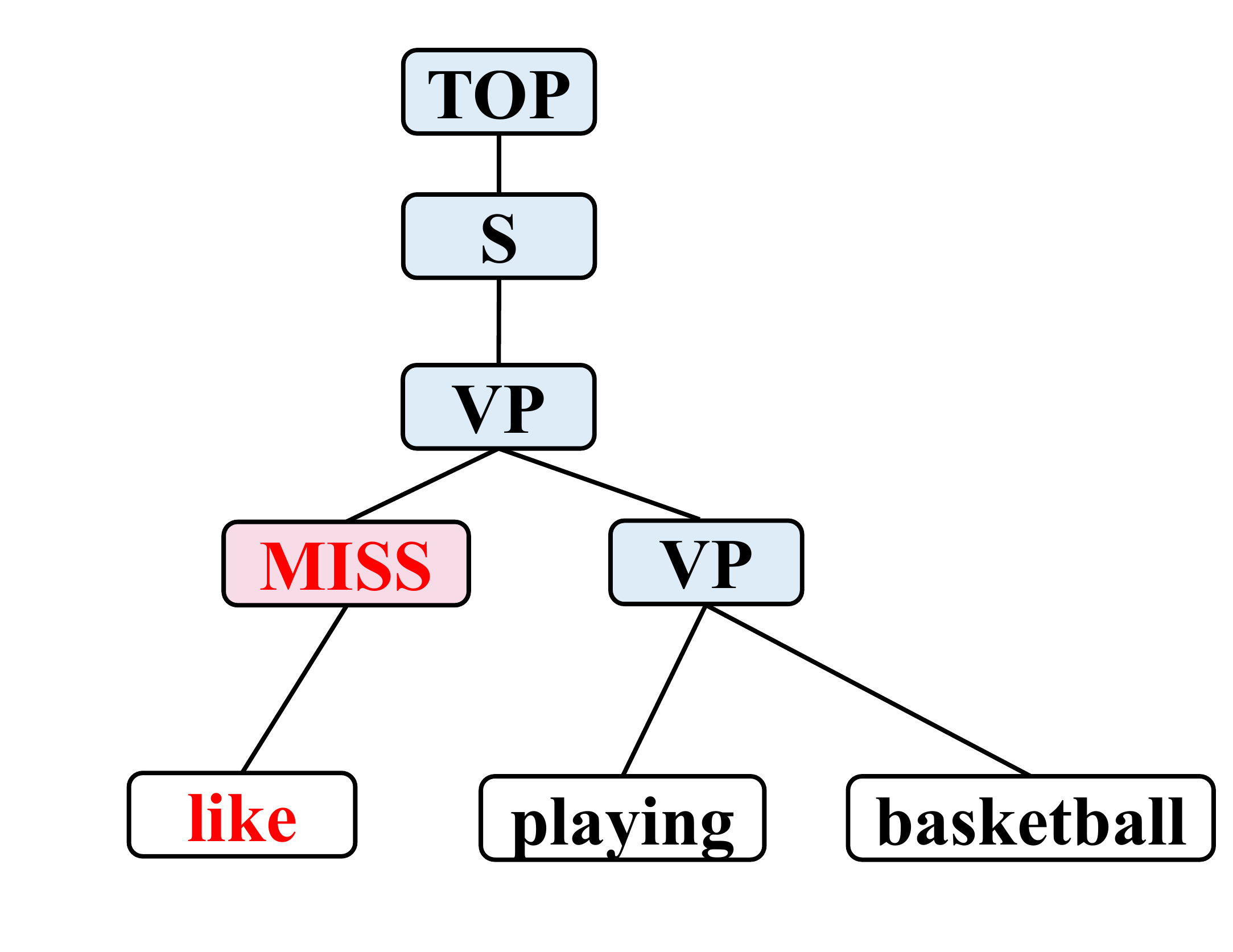}
    \caption{Missing errors.}
    \label{fig:m-error}
  \end{subfigure}
  \hfill
  \caption{Illustration of our extended syntax representation scheme for constituency parsing.}
  \label{fig:scheme-three-error-types}
\end{figure*}

Till now, the benefits of syntax in GEC have---for the most part---been demonstrated just within the
context of dependency-based syntax \cite{wan2021syntax, li2022syntax, zhang2022syngec}. Nevertheless, there is yet another kind of syntax, namely constituent-based syntax. It adopts hierarchical phrase-based trees to convey sentence structural information and has been proven to be useful in various downstream tasks \cite{DBLP:conf/emnlp/MarcheggianiT20,DBLP:conf/naacl/XiaZWLZHSZ21, DBLP:conf/naacl/ZhouLT22, bharadwaj-shevade-2022-efficient}.

\textcolor{black}{
This work first manages to explore whether---or to what extent---can constituent-based syntax contribute to GEC. After witnessing the success of SynGEC in exploiting dependency-based syntax, we extend it to constituent-based syntax, and propose CSynGEC (Constituency SynGEC).
We imitatively extend the constituent-based syntax scheme to accommodate grammatical errors and leverage parallel GEC data to train a GEC-oriented parser (GOPar) for constituency parsing.
To encode syntactic knowledge, we employ the graph convolutional network (GCN) \cite{DBLP:conf/iclr/KipfW17}.
Experimental results show that CSynGEC significantly improves the performance over the Transformer baseline, which proves that---with appropriate adaptation---constituent-based syntax is another useful external knowledge for GEC models.
}

\textcolor{black}{
Our work further attempts to investigate the integration of constituency-based and dependency-based syntax for GEC, as previous work has proven that they are complementary in helping downstream tasks \cite{li-etal-2010-combining, kong2011combining, fei-etal-2021-better}.
To this end, we experiment with two methods: 1) intra-model combination: we employ two separate GCNs to encode both kinds of syntax and fuse the representations for final decoding; 2) inter-model combination: we build multiple models enhanced with different kinds of syntax, and selecting edits predicted by them to achieve final results. We find that both methods improve the overall performance over just using a single kind of syntax, while their behaviors are different. Specifically, the intra-model method improves recall, but the inter-model method improves precision.
}

\section{GEC-Oriented Parser}
We first build a GEC-oriented parser (GOPar) to get the tailored constituent-based syntax of ungrammatical inputs.
\label{sec:par}

\textbf{Extended syntax representation scheme for constituency parsing.} As discussed in \citet{zhang2022syngec}, one stumbling block to parsing ungrammatical sentences is the inadaptability of the syntax representation scheme. More specifically, the original scheme may not represent the non-canonical structures arising from errors well. So, following \citet{zhang2022syngec}, we extend the syntax representation scheme for constituency parsing via several straightforward rules.

The basic idea is to insert pseudo non-terminal nodes\footnote{In constituency trees, terminal nodes are the words in sentences and non-terminal nodes are hierarchical constituents, such as ``VP'', ``NP'', ``SBAR'', etc.}  into the constituency tree to identify erroneous words. We design rules to handle \textit{substituted}, \textit{redundant}, and \textit{missing} errors, as discussed below.

\begin{enumerate}
    \item[(1)] \textbf{Substituted errors (SUB)} mean that erroneous words should be substituted with other words. For such errors, we insert a pseudo ``SUB'' non-terminal node as the new head of the erroneous word, as shown in Figure \ref{fig:s-error}.
    \item[(2)] \textbf{Redundant errors (RED)} mean that erroneous words should be deleted. We put a redundant word into the phrase to which its right-side word belongs, even if its right-side word is redundant also. If the redundant word is at the end of the sentence, we choose its left-side word instead. Then, we insert a pseudo ``RED'' non-terminal node as the new head of the redundant word, as shown in Figure \ref{fig:r-error}.
    \item[(3)] \textbf{Missing errors (MISS)} mean that some words should be inserted. For each missing word, we insert a pseudo ``MISS'' non-terminal node as the head of its right-side adjacent word, as shown in Figure \ref{fig:m-error}. We choose the left-side word if the missing word is at the end of the sentence. If multiple words are missed at the same position, we still only insert one ``MISS'' node.
\end{enumerate}

\textcolor{black}{\textbf{Training GOPar for constituency parsing.}} We take the idea from \citet{zhang2022syngec} to train a GOPar for constituency parsing using parallel GEC data as a pivot. Firstly, we parse the target-side correct sentences in GEC corpora by an off-the-shelf constituency parser. Secondly, we extract all errors in the source-side incorrect sentences through ERRANT\footnote{\url{https://github.com/chrisjbryant/errant}} \cite{bryant2017automatic}. Thirdly, we project the target-side constituency trees to the source-side ones by adjusting the structures of the erroneous part via the rules mentioned above and keeping the correct part unchanged. Finally, we utilize the projected trees to train GOPar.

\textbf{Discussion.} Compared with the dependency-based syntax scheme proposed by \citet{zhang2022syngec}, our scheme based on constituent-based syntax can represent grammatical errors more flexibly. Specifically, their scheme fails to handle the case that different errors occur at the same position, as multiple labels can not be assigned to one dependency arc. Instead, we can insert multiple non-terminal nodes for marking one word. Please kindly note that our scheme is lightweight and still has much room for improvement. 

\section{Syntax-Enhanced GEC Model}
\label{sec:model}

Our syntax-enhanced GEC model is built upon Transformer \cite{vaswani2017attention}. 
We utilize GCN \cite{DBLP:conf/iclr/KipfW17} to encode syntactic information and fuse the results with the outputs of Transformer encoder for decoding.

\textbf{Transformer backbone.} Transformer contains two parts: an encoder for encoding the source sentence and a decoder for predicting based on the output of the encoder and the generated tokens. For more details, please refer to \citet{vaswani2017attention}.

\textbf{GCN over constituency trees.} We employ GCN to encode  constituency trees following previous studies \cite{DBLP:conf/emnlp/MarcheggianiT20, DBLP:conf/naacl/XiaZWLZHSZ21, DBLP:conf/naacl/ZhouLT22}.
\textcolor{black}{
We view the constituency tree as an undirected graph. 
The nodes in this graph are all terminal/non-terminal nodes in the constituency tree, and edges are constructed according to the connections in the tree while ignoring the directions.
}
The output $h_v^{(l)}$ of the $l$-th layer GCN for node $v$ are computed as:

\begin{equation}
    \mathbf{h}_v^{(l)} = \mbox{ReLU}(\sum_{u\in\mathcal{N}(v)}W^{(l)}\mathbf{h}_u^{(l-1)}+\mathbf{b}^{(l)})
\end{equation}
where $\mathcal{N}(v)$ refers to the set of all one-hop neighbours of node $v$.
$W\in\mathbb{R}^{d \times d}$ and $\mathbf{b}\in\mathbb{R}^{d}$ are the learnable weight matrix and bias term.
ReLU \citep{nair2010rectified} is the activation function.
\textcolor{black}{For the initial input $\mathbf{h}_{v_t}^{(0)}$ of terminal node $v_t$ (i.e., the token), we directly use the outputs of the Transformer encoder. To derive the initial input $\mathbf{h}_{v_{nt}}^{(0)}$ of non-terminal node $v_{nt}$ (i.e., the constituent), 
we randomly
initialize a non-terminal embedding matrix $E_{nt}^{n \times d}$, where $n$ means the number of constituent labels and $d$ is the embedding dimension. The above node initialization method is borrowed from \citet{DBLP:conf/naacl/XiaZWLZHSZ21}, and still leaves room for improvement. For example, the non-terminal embedding matrix can be pre-trained with existing treebanks like PTB \cite{marcinkiewicz1994building}.}

\textbf{Representation fusion.} We use the weighted-sum operation to fuse syntax-enhanced representation $\mathbf{h}^{syn}$ and the outputs of the basic Transformer encoder $\mathbf{h}^{basic}$ as the final representation $\mathbf{h}^{final}$:
\begin{equation}
    \mathbf{h}^{final}=\lambda \mathbf{h}^{syn} + (1-\lambda) \mathbf{h}^{basic}
\end{equation}
where $\lambda \in (0,1)$ is the fusion factor. $\mathbf{h}^{basic}$ will later be fed into the decoder for token predicting.

\begin{figure}[tp!]
\centering
\includegraphics[scale=0.23]{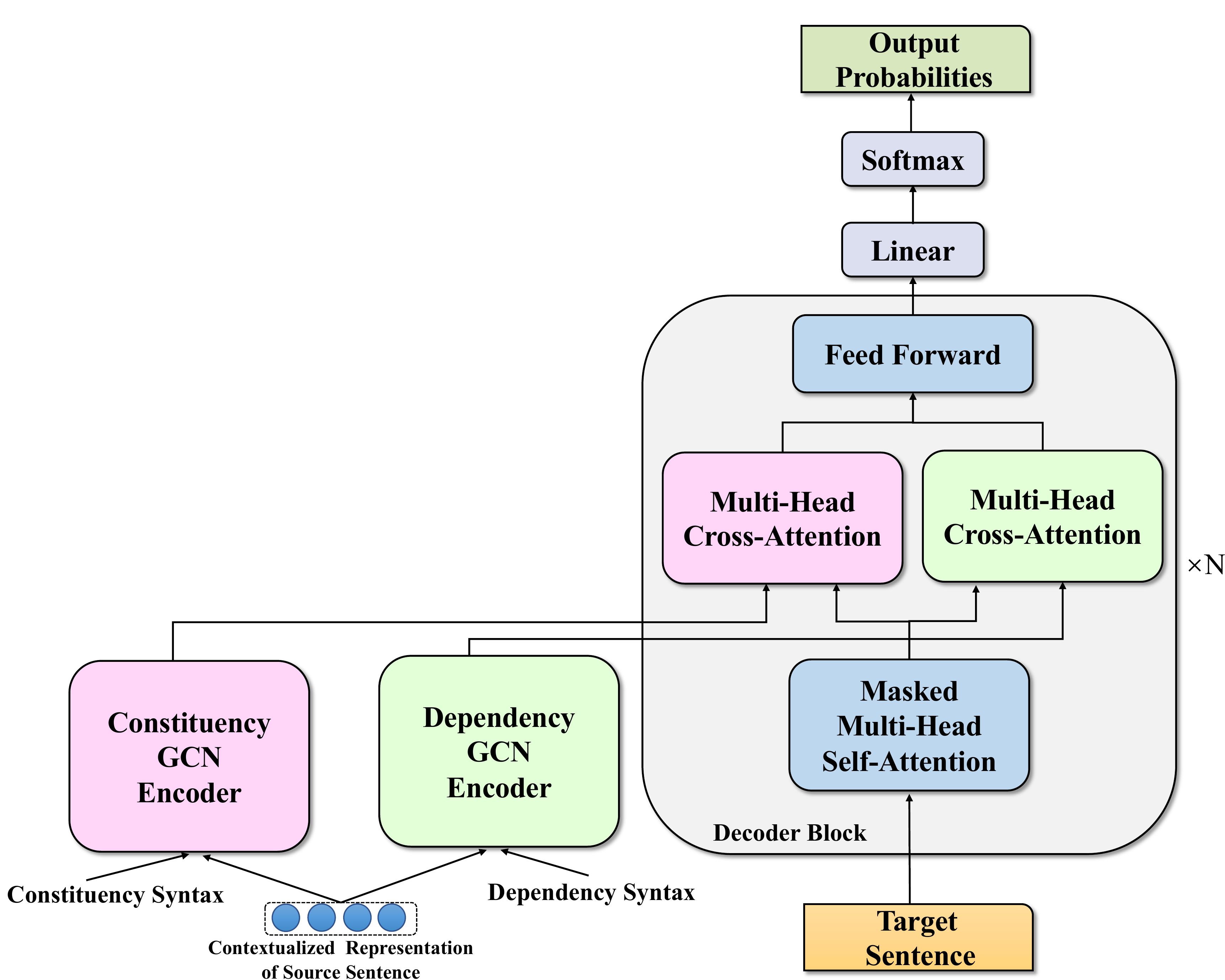}
\caption{Illustration of intra-model combination.}
\label{fig:combine}
\end{figure}

\section{Syntax Combination}
\label{sec:3.4}
\textcolor{black}{We present two methods to combine constituent-based syntax from this work and dependency-based syntax from \citet{zhang2022syngec} to help GEC.}

\textcolor{black}{\textbf{Intra-model combination.} This method integrates two syntax formalisms within a single Transformer. We first build two separate GCNs to encode two kinds of syntactic information. Then, we leverage two multi-head cross-attention layers to attend to them in the decoder. The results are added together for subsequent calculation. The whole procedure is depicted in Figure \ref{fig:combine}.}

\textcolor{black}{\textbf{Inter-model combination.} This method ensembles independent models enhanced with different kinds of syntax. We directly adopt the ensemble technique proposed by \citet{qorib-etal-2022-frustratingly}. We train multiple models based on either one kind of syntax, and gather all edits predicted by those models. Then, we use logistic regression to predict whether each edit should be retrained or discarded, and re-apply the preserved edits to get the final results.}

\section{Experiments}
\subsection{Experimental Setup}
Following previous work \cite{zhang2022syngec}, we use the cleaned version of the Lang8 dataset (CLang8) \cite{rothe2021recipe}, the FCE dataset \cite{yannakoudakis2011new}, the NUCLE dataset \cite{dahlmeier2013building} and the WI+LOCNESS train-set \cite{bryant2019bea} for training. We use the  BEA-19-\textit{Dev} dataset \cite{bryant2019bea} for validating. We report average (P)recision, (R)ecall, and (F$_{0.5}$) metrics on the CoNLL-14-\textit{Test} \cite{ng2014conll} and BEA-19-\textit{Test} \cite{bryant2019bea} datasets with their official evaluation tools \cite{dahlmeier2012better,bryant2017automatic} with 3 random seeds.
We present more data statistics and implementation details in Appendix \ref{sec:app:a}.

\begin{table}[tp!]
\centering
\scalebox{0.6}{
\begin{tabular}{lccc}
\toprule
                                   &\textbf{Extra}     & \textbf{CoNLL-14-\textit{test}}     & \textbf{BEA-19-\textit{test}}     \\
                                 \textbf{Model}& \textbf{Data Size}  & \textbf{P/R/$\mbox{\textbf{F}}_{0.5}$}    & \textbf{P/R/$\mbox{\textbf{F}}_{0.5}$} \\ \hline
                                  \multicolumn{4}{c}{\textbf{w/o PLM}} \\ \hline
                                  \multicolumn{4}{l}{\textbf{w/o syntax}}     \\
                                   \citet{kiyono2019empirical} &   70M         & 67.9/44.1/61.3          & 65.5/59.4/64.2          \\
                                  \citet{lichtarge-etal-2020-data}&   340M        & 69.4/43.9/62.1          & 67.6/62.5/66.5          \\
                                  \citet{stahlberg2021synthetic}&   540M     & 72.8/49.5/\textbf{66.6}          & 72.1/64.4/\textbf{70.4}        \\
                                  \textbf{Our Baseline} &   2.4M &  66.9/40.3/59.1   & 66.8/55.5/64.2         \\
                                  \hdashline 
                                  \multicolumn{4}{l}{\textbf{w/ syntax}} \\ 
                                   \citet{wan2021syntax} &   10M       & 74.4/39.5/63.2          & 74.5/48.6/67.3          \\
                                 \citet{li2022syntax} &   30M      & 66.7/38.3/58.1          & -/-/-          \\
                                  \textbf{DSynGEC} \cite{zhang2022syngec} &   2.4M  & 70.0/46.2/\textbf{63.5}          & 70.9/59.9/\textbf{68.4}          \\
                                    \textbf{CSynGEC} (this work) &   2.4M  & 69.7/46.3/63.3          & 69.4/60.3/67.4 
                                  \\\hline \hline
                                     \multicolumn{4}{c}{\textbf{w/ PLM}} \\ \hline 
                                     \citet{DBLP:conf/acl/SunGWW20}&   300M          & 71.0/52.8/66.4       & 74.7/66.4/72.9          \\
                                      \citet{rothe2021recipe}$^{*}$ &   2.4M         & -/-/\textbf{68.8}          & -/-/\textbf{75.9}          \\  \textbf{Our Baseline} &   2.4M &  73.6/48.6/66.7   & 74.0/64.9/72.0         \\
                                       \textbf{DSynGEC} \cite{zhang2022syngec} &   2.4M  & 74.7/49.0/67.6          & 75.1/65.5/72.9          \\
                                       \textbf{CSynGEC} (this work) &   2.4M  & 74.0/50.7/67.7         & 74.4/66.1/72.6          \\

\bottomrule
\end{tabular}
}
\caption{\textbf{Single-model} results. ``\textbf{w/ syntax}'' means using syntactic knowledge. ``\textbf{w/ PLM}'' means using pre-trained language models. $^{*}$ denotes current SOTA.}

\label{tab:main:results}
\end{table}

\subsection{Experimental Results}
\textbf{Main results} are shown in Table \ref{tab:main:results}.
\textcolor{black}{When not using pre-trained language models (w/o PLM), CSynGEC gets 63.3 and 67.4 F$_{0.5}$ scores on CoNLL-14-\textit{Test} and BEA-19-\textit{Test}, respectively, which is quite competitive. Compared with the baseline, incorporating tailored constituent-based syntax leads to significant F$0.5$ gains (+4.2/+3.2) on both benchmarks, which clearly shows its effectiveness. We also build a stronger baseline by initializing the parameters of the Transformer backbone from BART \cite{lewis2020bart} (w/ PLM). CSynGEC still achieves +1.0/+0.6 F$_{0.5}$ improvements over BART, which indicates that the effectiveness of our method will not easily be overwhelmed by PLMs.}

To distinguish, we rename the method in \citet{zhang2022syngec} as DSynGEC (Dependency SynGEC). We observe that CSynGEC outperforms all other syntax-enhanced counterparts, except DSynGEC. 
\textcolor{black}{Specifically, we find that DSynGEC always performs better in precision, while CSynGEC always achieves higher recall. Such a discrepancy further motivates us to combine them.}

The SOTA results are kept by \citet{rothe2021recipe}, which leverages a huge PLM with up to 11B parameters\footnote{While CSynGEC only has 70M parameters.}. Since CSynGEC is model-agnostic, we will test it in more SOTA baselines in the future.

\begin{table}[tp!]
\centering
\scalebox{0.65}{
\begin{tabular}{lcc}
\toprule
                                     & \textbf{CoNLL-14-\textit{test}}     & \textbf{BEA-19-\textit{test}}     \\
                                 \textbf{Model}  & \textbf{P/R/$\mbox{\textbf{F}}_{0.5}$}    & \textbf{P/R/$\mbox{\textbf{F}}_{0.5}$} \\ \hline
                                 \textbf{Baseline}        & 66.7/40.3/59.1        & 66.8/55.5/64.2         \\ \hdashline
                                 \textbf{DSynGEC \cite{zhang2022syngec}}        &\textbf{70.0}/46.2/\textbf{63.5}         & \textbf{70.9}/59.9/\textbf{68.4}          \\ 
                                   \textbf{CSynGEC}        & 69.9\textbf{/46.3/}63.3       & 69.4\textbf{/60.3/}67.4        \\ 
                                   \hspace{0.3cm} \textbf{w/ off-the-shelf parser}        & 66.6/41.4/59.4          & 67.0/56.4/64.6          \\
                                   \hspace{0.3cm} \textbf{w/o extended scheme}        & 68.4/43.2/61.3          & 67.6/56.7/65.1          \\ \hdashline
                                   \multicolumn{3}{c}{\textit{\textbf{intra-model combination}}} \\ 
                                         \textbf{CSynGEC + DSynGEC}        & \textbf{69.6/47.7/63.8}          & \textbf{70.4/61.7/68.5}          \\
                                   \hspace{0.3cm} \textbf{w/o independent cross-attention}        & 67.6/47.5/62.3          & 68.5/61.6/67.0          \\\hdashline
                                   \multicolumn{3}{c}{\textit{\textbf{inter-model combination}}} \\ 
                                         \textbf{6 $\times$ DSynGEC}        & 77.0/39.4/64.7          & 81.2/54.7/74.1          \\
                                             \textbf{6 $\times$ CSynGEC}        & 77.7/\textbf{40.7}/65.7         & 81.8/\textbf{55.0}/74.5          \\
                                             \textbf{3 $\times$ DSynGEC + 3 $\times$ CSynGEC}        & \textbf{79.0}/40.1/\textbf{66.2}          & \textbf{83.0}/54.6/\textbf{75.2 }         \\
                                
\bottomrule
\end{tabular}
}
\caption{Results of model ablation and ensemble. }

\label{tab:abb}
\end{table}

\textbf{Benefits of GOPar.} We first study the effectiveness of using syntax derived from GOPar. We replace GOPar with an off-the-shelf constituency parser trained on the PTB treebank \cite{marcinkiewicz1994building}. As shown in Table \ref{tab:abb}, the performance of CSynGEC dramatically drops after changing the parser (-3.9/-2.8), which confirms the necessity of GOPar. We conjecture that off-the-shelf parsers tend to provide low-quality parses for ungrammatical sentences, which may introduce noise for GEC.

\textbf{Impact of extended syntax scheme.} We extend the constituent-based syntax scheme by inserting pseudo non-terminal nodes to represent grammatical errors. To study the impact of this task-oriented adaptation, we remove all inserted pseudo nodes before feeding the trees produced by GOPar into GEC models. From Table \ref{tab:abb}, we can see that CSynGEC heavily degenerates without the extended scheme (-2.0/-2.3). However, the performance is still better than using an off-the-shelf parser, as the high-quality parses for the correct part are kept.

\textcolor{black}{\textbf{Results of syntax combination.} From Table \ref{tab:abb}, we can see that the intra-model combination significantly improves recall over DSynGEC (+1.5/+1.8) and CSynGEC (+1.4/+1.4), and achieves better overall performance on both test sets. Besides, we note that this method will not work if we use a sharing cross-attention layer to attend to different kinds of syntactic information.}

\textcolor{black}{Under the inter-model combination setting, ``3 $\times$ DSynGEC + 3 $\times$ CSynGEC'' substantially improves precision over ``6 $\times$ DSynGEC'' (+2.0/+1.8) and ``6 $\times$ CSynGEC'' (+1.3/+1.2), and achieves the best F$_{0.5}$ scores. All these results demonstrate that constituent-based and dependency-based syntax has intrinsic complementary strength for helping GEC models. We also give a case in Appendix \ref{sec:case} to show how both kinds of syntax work.}

\section{Conclusion}
In this paper, we extend the idea of SynGEC \cite{zhang2022syngec} and propose the CSynGEC approach to enhance GEC models by exploiting tailored constituent-based syntax. Experimental results show that incorporating constituent-based syntax produced by a GEC-oriented constituency parser can effectively help GEC models. 
Furthermore, we attempt to combine dependency-based and constituent-based syntax from both intra-model and inter-model aspects, and find that simultaneously using two kinds of syntax leads to more obvious improvement.

\bibliography{anthology,custom}
\bibliographystyle{acl_natbib}

\appendix
\vspace{+1ex}
\begin{center}
\Large \textbf{Appendices} 
\end{center}

\vspace{+1ex}

\section{Experimental Details}
\begin{table}[h!]
\scalebox{0.73}{
\begin{tabular}{lccc}
\toprule
\textbf{Dataset} & \textbf{\#Sentences} & \textbf{\%Error} & \textbf{Usage}    \\ \hline
\textbf{CLang8}            & 2,372,119            & 57.8             & Pre-training         \\ 
\textbf{FCE}              & 34,490               & 62.6             & Fine-tuning I     \\
\textbf{NUCLE}            & 57,151               & 38.2             & Fine-tuning I     \\
\textbf{W\&I+LOCNESS}     & 34,308               & 66.3             & Fine-tuning I\&II \\
\hline
\textbf{BEA-19-\textit{Dev}}              & 4,384               & 65.2             & Validation     \\
\textbf{CoNLL-14-\textit{Test}}              & 1,312               & 72.3             & Testing     \\
\textbf{BEA-19-\textit{Test}}             & 4,477               & N/A             & Testing     \\ 
\bottomrule
\end{tabular}
}
\caption{Statistics of used datasets. \textbf{\#Sentences} denotes the total number of sentences. \textbf{\%Error} refers to the proportion of erroneous sentences.}
\label{tab:dataset}
\end{table}
\label{sec:app:a}
The statistics of all datasets used in this work are shown in Table \ref{tab:dataset}, please refer to their original papers for corresponding download links.

\begin{figure}[h!]
\centering
\includegraphics[scale=0.38]{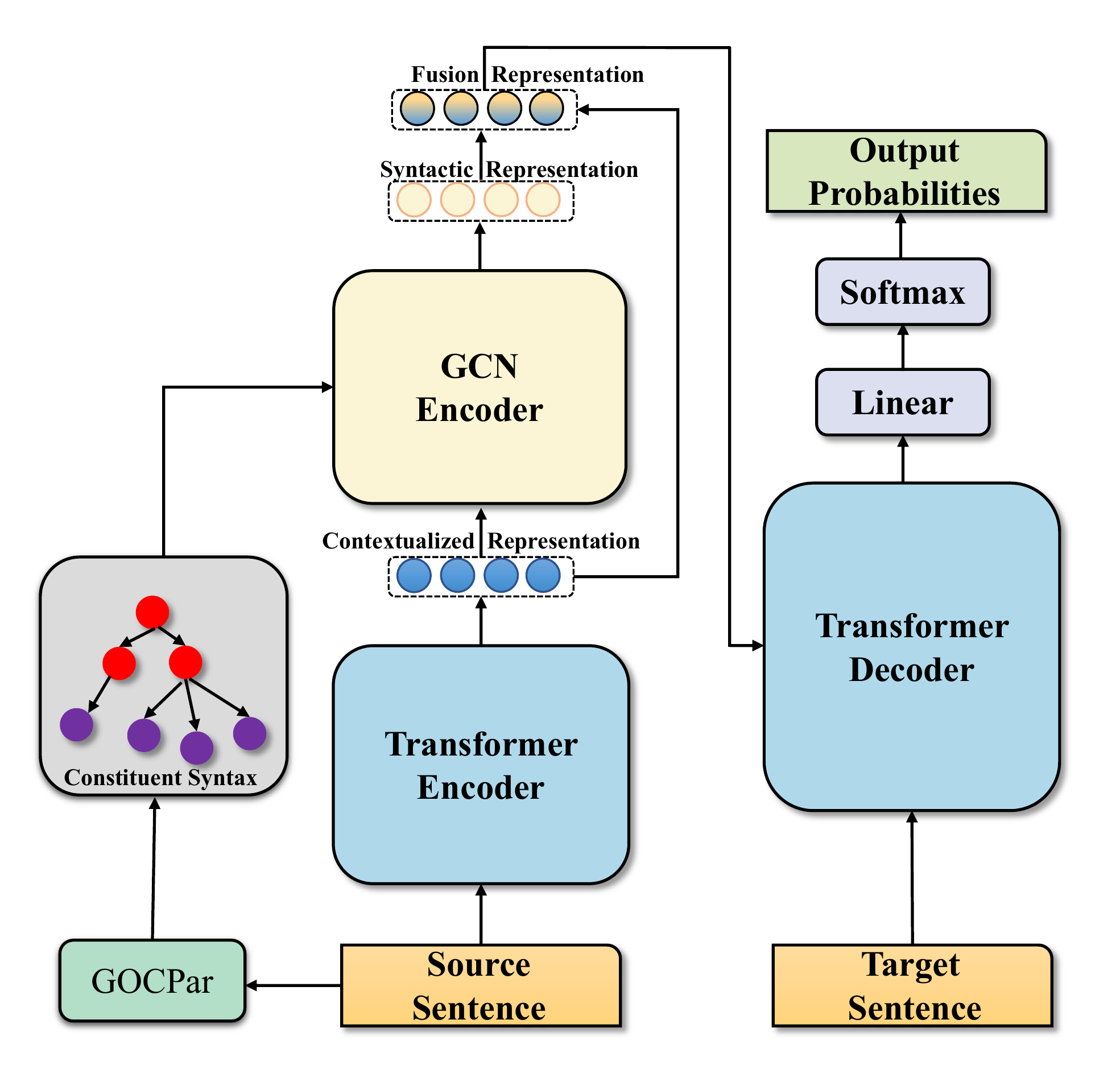}
\caption{The overview of our syntax-enhanced GEC model.}
\label{fig:model}
\end{figure}
For the GEC model, we build it with the \texttt{Fairseq}\footnote{\url{https://github.com/facebookresearch/fairseq}} toolkit \cite{ott2019fairseq}. The backbone model is a Transformer with the ``base'' configuration. To encode syntactic information, 3 GCN blocks are stacked. Other hyper-parameters and training tricks are directly taken from \citet{zhang2022syngec}. The architecture of our syntax-enhanced GEC model is depicted in Figure \ref{fig:model}.

For GOPar, we adopt the \texttt{Supar}\footnote{\url{https://github.com/yzhangcs/parser}} toolkit and follow the default settings. We employ ELECTRA \cite{DBLP:conf/iclr/ClarkLLM20} as its encoder. 
The training data is automatically derived from the CLang8 GEC dataset via tree projection.

\begin{figure}[h!]
\centering
\includegraphics[scale=0.17]{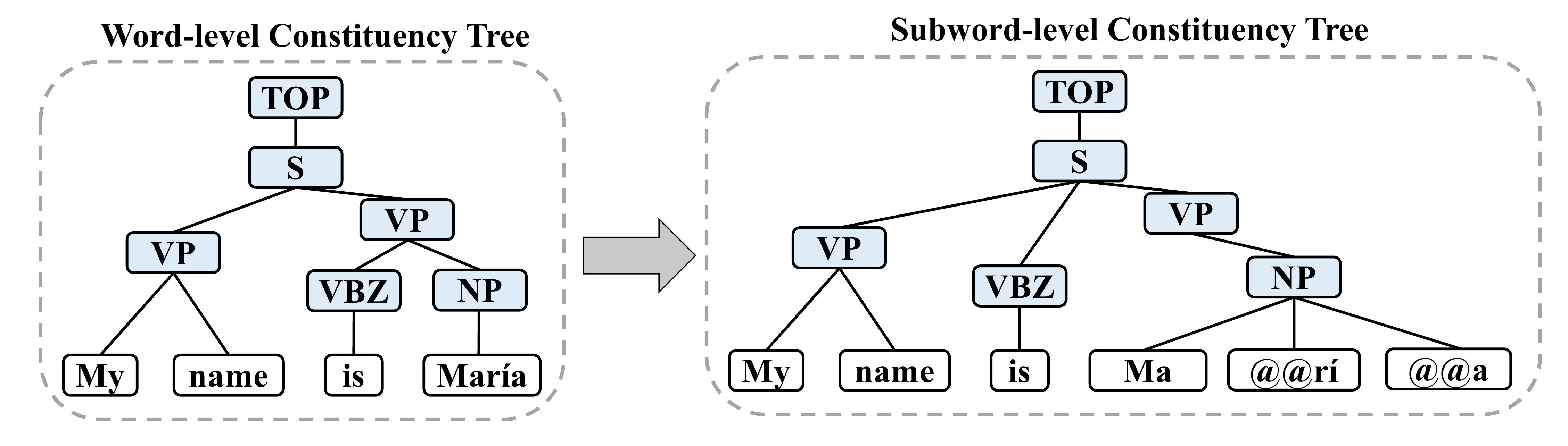}
\caption{The illustration of converting word-level trees to subword-level trees.}
\label{fig:sub}
\end{figure}

To fill the gap between the word-level parsing results and the subword-level Transformer inputs, we assign the head non-terminal node of the original word to all internal subwords, thus creating a subword-level tree, as shown in Figure \ref{fig:sub}. Then, we feed such subword-level trees into our syntax-enhanced GEC models.

\section{Case Study}

\begin{figure}[h!]
\centering
\includegraphics[scale=0.38]{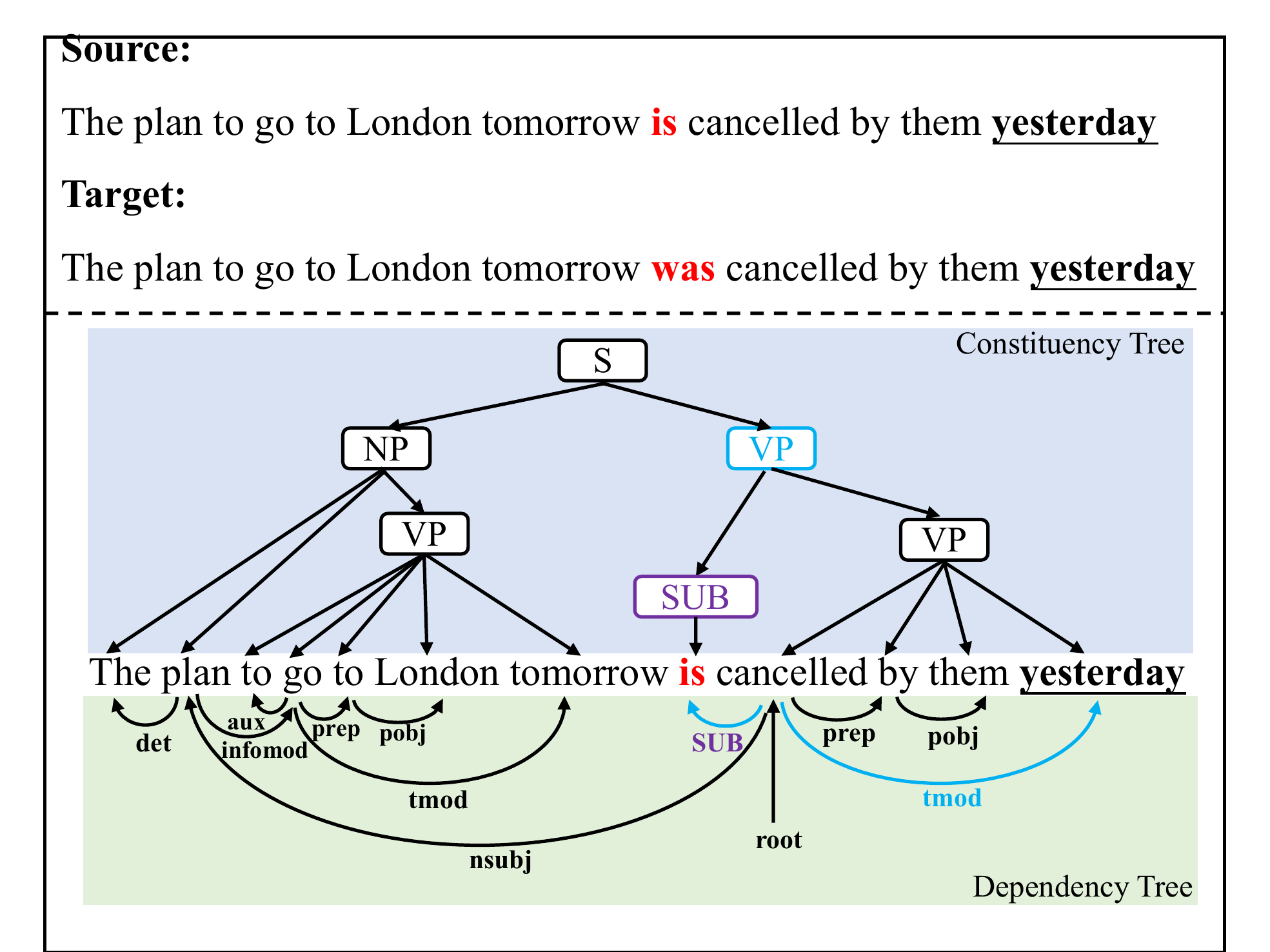}
\caption{A GEC example along with the tailored constituency and dependency trees of the source sentence.
}
\label{fig:example}
\end{figure}
\label{sec:case}
\textcolor{black}{A case study is presented in Figure \ref{fig:example}. 
There is a verb tense error in the source sentence and we should substitute ``is'' with ``was'' to correct it. The correction evidence for this error is ``yesterday'', which drops a hint that the past tense should be used here.}

\textcolor{black}{The tailored constituency tree from this work (in the top) indicates that the erroneous word ``is'' and the correction evidence ``yesterday'' are in the same ``VP'' phrase, while ``is'' and the noise information ``tomorrow'' are not (ignore the whole sentence constituent ``S''). What is more, the inserted pseudo non-terminal node ``SUB'' directly points out that ``is'' is a substituted error.}

\textcolor{black}{The tailored dependency tree from \citet{zhang2022syngec} (in the bottom) shows that ``is'' and ``yesterday'' are siblings and have the same father node ``cancelled'', and ``yesterday'' is a time modifier (``tmod''). The label ``SUB'' of the incoming arc of ``is'' also supports that ``is'' should be substituted.}

\textcolor{black}{In short, both kinds of syntax provide helpful clues for GEC from different perspectives.}

\end{CJK}
\end{document}